\documentclass[letterpaper]{article}

\usepackage{natbib,alifeconf}  %% The order is important
\usepackage{amsmath,amssymb}   %% math: \text, \tfrac, \bigl (used in the Propositions)
\usepackage{multirow}          %% \multirow in the assignment-method table
\usepackage{url,hyperref,cleveref}
\usepackage{booktabs}

\usepackage{lipsum}

\newcommand\blfootnote[1]{%
  \begingroup
  \renewcommand\thefootnote{}\footnote{#1}%
  \addtocounter{footnote}{-1}%
  \endgroup
}

\title{Per-Node Activation Function Evolution in Indirectly Encoded Substrates: Solvability, Limits, and Emergent Diversity}

\author{
    Romain Claret$^{1}$,
    Michael O'Neill$^{2}$,
    Paul Cotofrei$^{1}$, \and
    Kilian Stoffel$^{1}$ \\
    \mbox{}\\
    $^1$University of Neuch\^atel, Switzerland \\
    $^2$University College Dublin, Ireland
    romain.claret@unine.ch
} % email of corresponding author

\newtheorem{proposition}{Proposition}

\newcommand{\hmr}{EMR-HyperNEAT}

\newcommand{\cppn}{CPPN}

\begin{document}

\maketitle

%% =============================================================================
%% ABSTRACT
%% =============================================================================
\begin{abstract}
Biological neurons achieve computational diversity through specialized types: tonic, bursting, adapting, and fast-spiking cells coexist within the same circuit. Artificial neural networks, by contrast, apply a single activation function uniformly to all nodes, which limits what they can represent. We show that this uniformity creates hard limits for evolutionary search: across sparse evolved substrates, monotonic functions fail to solve parity beyond its smallest instance, XOR, while a single oscillatory unit suffices at all tested scales. The gap is one of search and sparsity, not representation: monotonic networks can represent parity with a modest number of hidden units, and gradient descent recovers that solution. We evolve, to our knowledge for the first time in indirect encoding, per-node activation function assignments from an 18-function palette across more than 4{,}500 experimental runs spanning Boolean logic, regression, and spatial classification.

Testing each of the 18 functions individually on Parity-4 reveals a three-tier solvability structure: oscillatory functions achieve 100\%, intermediate functions 6.7--80\%, and all 9 monotonic functions 0\%. This divide is not universal. Recurrence collapses it, and gradient descent inverts it entirely, showing that the barrier is specific to evolutionary search in sparse substrates. What activation functions a network can use, beyond its topology and weights, determines what evolutionary search can solve. Indirect encoding discovers heterogeneous per-node activation assignments unlikely to be chosen by hand.
\end{abstract}

Data/Code available at: \url{https://github.com/RomainClaret/emr-hyperneat}
\blfootnote{\textcopyright{} 2026 Romain Claret, Michael O'Neill, Paul Cotofrei, Kilian Stoffel. Published under a Creative Commons Attribution 4.0 International (CC BY 4.0) license.}

%% =============================================================================
%% SECTION 1: INTRODUCTION
%% =============================================================================
\section{Introduction}
\label{sec:introduction}

Every cortical circuit contains a mix of neuron types with distinct firing properties~\citep{izhikevich2003simple,markram2004interneurons}, yet most artificial neural networks force every node to use the same activation function~\citep{lecun2015deep}. This uniformity is consequential. It imposes hard limits on evolutionary search: although monotonic functions like tanh can represent parity given enough width, neuroevolution in sparse substrates fails to discover such solutions, while a single oscillatory unit suffices. We formalize this gap in this paper.

Can indirect encoding evolve per-node activation function assignments that match computational substrates to problem structure? This question matters because activation function selection is not hyperparameter optimization: it determines what a network \textit{can} represent. Choosing the wrong function category makes a problem unsolvable; choosing the right one can make it trivial.

Why use evolutionary search rather than hand-designed topologies? Assigning heterogeneous activation functions per node creates a combinatorial space that grows exponentially with network size. Indirect encoding via a Compositional Pattern Producing Network (\cppn{})~\citep{stanley2007compositional} makes this search tractable: the CPPN generates the substrate topology and connection weights from spatial coordinates, which we extend to also assign each node an activation function, so that compact genomes can produce large structured networks with heterogeneous functions. Evolution discovers function assignments unlikely to be chosen by hand.

We test three assignment methods (CPPN-output, weight-interpretation, uniform) across 18 activation functions spanning 7 categories, and compare evolutionary search with gradient-based optimization, revealing that the solvability divide reverses under gradient descent. We address three research questions:

\begin{itemize}
    \item \textbf{RQ1}: Can neuroevolution with indirect encoding discover per-node activation assignments, and does the choice of function category create hard solvability limits?
    \item \textbf{RQ2}: Which assignment mechanism is most effective for per-node function selection via CPPN?
    \item \textbf{RQ3}: Is the solvability divide universal, or specific to particular problem structures and optimizers?
\end{itemize}

\subsection*{Contributions}

\begin{enumerate}
    \item \textbf{Per-function solvability characterization} (RQ1): We individually test all 18 activation functions from the palette on Parity-4, revealing a three-tier solvability structure (100\% / 6.7--80\% / 0\%) quantitatively predicted by local extrema count (Spearman $\rho{=}0.81$). We separate representability from discoverability: monotonic networks can represent parity with width $\Theta(n)$, yet neuroevolution does not find it in sparse substrates, while a single oscillatory unit suffices (Propositions~\ref{prop:oscillatory},~\ref{prop:monotonic}).
    \item \textbf{Assignment method and palette robustness} (RQ2): CPPN-controlled per-node assignment significantly outperforms heuristic and uniform methods, larger palettes slow convergence without reducing success, and the divide persists across population sizes 50--2{,}000.
    \item \textbf{Boundary conditions} (RQ3): Recurrence collapses the divide (monotonic 0\%$\rightarrow$80--100\% on Parity-4), gradient descent inverts it (tanh maintains 100\% through Parity-8), and the divide is absent on non-parity tasks (all 18 functions solve sine regression at 100\%). The divide is specific to evolutionary search in indirect encoding, not universal.
\end{enumerate}

%% =============================================================================
%% SECTION 2: BACKGROUND
%% =============================================================================
\section{Background}
\label{sec:background}

\subsection{Indirect Encoding and HyperNEAT}

NEAT~\citep{stanley2002evolving,papavasileiou2021systematic} is a direct encoding that evolves neural network topology and weights simultaneously through speciation and incremental complexification. HyperNEAT~\citep{stanley2009hypercube} extends NEAT with indirect encoding: NEAT evolves a compact \cppn{}~\citep{stanley2007compositional} that generates connection weights from spatial coordinates $w_{ij} = \text{CPPN}(x_i, y_i, x_j, y_j)$, exploiting geometric regularities~\citep{gauci2008case} to produce large structured networks from small genomes. ES-HyperNEAT~\citep{risi2012enhanced} adds evolvable substrates via variance-based neuron placement. \hmr{}~\citep{claret2026emr} replaces ES-HyperNEAT's sequential quadtree with a dense multi-resolution grid evaluated eagerly as tensor operations, for GPU acceleration.

Standard HyperNEAT applies uniform activation (tanh) to all substrate nodes. We extend this to allow per-node activation selection from a predefined palette of 18 functions.

\subsection{Activation Function Diversity}

Activation function handling in the NEAT lineage has progressed through three levels.
\textit{Level~1 (fixed)}: Original NEAT~\citep{stanley2002evolving} applies a steepened
sigmoid to every node; no per-node diversity exists.
\textit{Level~2 (random-at-creation, CPPN only)}: HyperNEAT~\citep{stanley2009hypercube} assigns each \textit{CPPN} node a random
activation function at creation, but the decoded substrate uses uniform activation.
\textit{Level~3 (mutable per-node)}: HA-NEAT~\citep{hagg2017evolving} and
HAFD-NEAT~\citep{papavasileiou2022hafdneat} evolve per-node activations via direct
encoding. We bring Level~3 to indirect encoding: a CPPN output
assigns activation functions to substrate nodes from spatial coordinates, which produces
spatially structured functional regions without per-gene mutation.

We are not aware of prior work that evolves per-node substrate activation via indirect encoding. We also give a formal account of both sides of the oscillatory/monotonic divide and characterize all 18 functions individually.

%% =============================================================================
%% SECTION 3: RELATED WORK
%% =============================================================================
\section{Related Work}
\label{sec:related}

\paragraph{Heterogeneous activations.}
All prior per-node activation work operates in direct
encoding~\citep{hagg2017evolving,papavasileiou2022hafdneat}. The indirect-encoding
mechanism presented here produces spatially structured functional regions (described in the Methods). Our results
extend Hagg et al.'s finding that mixed activations improve parsimony: we show function
\textit{category} creates a categorical solvability divide that goes beyond a size advantage.
Prior gradient-based approaches (PANGAEA~\citep{bingham2022pangaea}, Ramachandran et al.~\citep{ramachandran2017searching}) search for a single best function. We evolve per-node heterogeneous assignments with formal solvability analysis.

\paragraph{Periodic activations.}
SIREN~\citep{sitzmann2020siren} shows that periodic sinusoidal activations can enable implicit representations inaccessible to ReLU. Ziyin et al.~\citep{ziyin2020neural} formally show standard networks fail to learn periodic functions. Rahaman et al.~\citep{rahaman2019spectral} identify spectral bias: deep ReLU networks learn low-frequency components first. The Growing Cosine Unit~\citep{noel2021gcu} shows that a single oscillatory neuron solves XOR by creating multiple decision boundaries, in line with our Proposition~\ref{prop:oscillatory} (presented in the Formal Analysis). Sope\~{n}a et al.~\citep{sopena1999periodic} show that periodic perceptrons can compute arbitrary Boolean functions. Our Propositions~\ref{prop:oscillatory} and~\ref{prop:monotonic} formalize both directions of the divide.

\paragraph{Architecture search.}
NAS~\citep{zoph2016neural,real2019regularized} and CoDeepNEAT~\citep{miikkulainen2024evolving} search architectures including activations at the layer level. Nader and Azar~\citep{nader2021evolution} survey evolved activation functions. We instead assign activation functions per node, via indirect encoding.

%% =============================================================================
%% SECTION 4: METHODS
%% =============================================================================
\section{Methods}
\label{sec:methods}

\subsection{System Pipeline}

\begin{figure}[t]
\centering
\includegraphics[width=0.91\columnwidth]{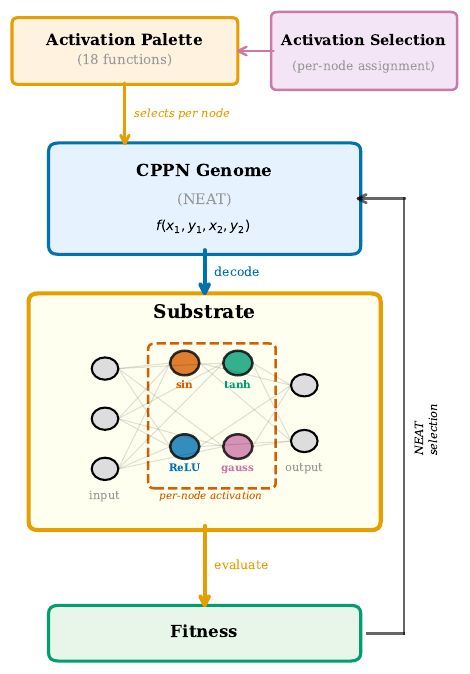}
\caption{System pipeline. The activation palette, through the per-node selection mechanism, assigns each node its own activation function; the CPPN genome (NEAT) is decoded into a substrate carrying these per-node activations, the substrate is evaluated for fitness, and NEAT selection closes the evolutionary loop. Three assignment methods determine how CPPN output maps to function selection.}
\label{fig:system_overview}
\end{figure}

Figure~\ref{fig:system_overview} shows the evolutionary pipeline. Each individual encodes a CPPN genome that generates a substrate by querying connection weights from spatial coordinates. The CPPN also assigns a per-node activation function from the palette: each substrate node receives its own activation function. The substrate is evaluated on the target problem, and standard NEAT selection produces the next generation.

\subsection{Three Per-Node Assignment Methods}
\label{sec:assignment_methods}

\paragraph{CPPN-output (Method A).} The CPPN produces an extra output channel beyond connection weights. The continuous value $x$ is squashed by $\tanh$ and binned into one of the $k$ palette indices: \mbox{$\text{idx} = \lfloor (\tanh(x){+}1)\,k/2 \rfloor$} (clipped to $[0,k{-}1]$). Spatial patterns in the CPPN naturally create structured functional regions.

\paragraph{Weight-interpretation (Method B).} Activation function indices are derived from incoming weight statistics without additional CPPN outputs, mapping a continuous statistic to a palette index (not a hard threshold). Three variants:
\begin{itemize}\setlength{\itemsep}{0pt}\setlength{\parskip}{0pt}
\item \textit{sign-based}: positive weights $\rightarrow$ tanh, negative $\rightarrow$ relu, mixed $\rightarrow$ sigmoid;
\item \textit{magnitude-based}: larger mean $|w|$ $\rightarrow$ bounded;
\item \textit{variance-based}: larger weight variance $\rightarrow$ nonlinear, e.g.\ $\tanh(5\,\mathrm{Var}[w])$.
\end{itemize}

\paragraph{Uniform (Method C).} All substrate nodes receive the same activation function. This serves as the baseline.

\subsection{Function Palette: 18 Activation Functions}
\label{sec:palette}

Table~\ref{tab:palette} lists all 18 functions organized into 7 categories. The palette spans oscillatory functions (which cross decision thresholds multiple times), monotonic functions (which cross at most once), and intermediate categories.

\begin{table}[htbp]
\centering
\caption{18 activation functions and their individual Parity-4 results (\protect\nameref{sec:ablation}; $N{=}30$ seeds; 300 gen max). The solvability boundary follows oscillatory structure: functions with local extrema solve parity; 9 strictly monotonic functions never solve.}
\label{tab:palette}
{\scriptsize\setlength{\tabcolsep}{2pt}
\begin{tabular}{rlllcc}
\toprule
Idx & Name & Formula & Category & Solve\% & Avg Gen \\
\midrule
\multicolumn{6}{l}{\textit{Multi-crossing} (100\% solve, 2.9--69.2 gen)} \\[2pt]
4 & sin & $\sin(x)$ & Oscillatory & \textbf{100} & 2.9 \\
13 & osc\_adapt & $\sin(x)(1{-}0.2|x|)$ & Oscillatory & \textbf{100} & 3.7 \\
11 & burst & $\tanh(x){+}0.5\sin(3x)$ & Oscillatory & \textbf{100} & 10.7 \\
16 & band\_pass & $e^{-|x-1|}{-}e^{-|x+1|}$ & Adaptive & \textbf{100} & 69.2 \\
\midrule
\multicolumn{6}{l}{\textit{Intermediate} (6.7--80\% solve, 55--246 gen)} \\[2pt]
15 & receptive & $e^{-x^2}\cos(2x)$ & Quasi-Osc. & 80 & 100.1 \\
12 & resonator & $\sin(x)\,e^{-|x|/3}$ & Oscillatory & 66.7 & 55.9 \\
5 & gauss & $e^{-x^2}$ & Radial & 33.3 & 139.3 \\
8 & rs\_adapt & $\tanh(x)(1{-}0.3|x|)$ & Adaptive & 23.3 & 202.4 \\
10 & lts\_low & $\sigma(2x{-}0.5)$ & Mono./Bounded & 6.7 & 246.0 \\
\midrule
\multicolumn{6}{l}{\textit{Monotonic} (0\% solve, 9 functions)} \\[2pt]
0 & tanh & $\tanh(x)$ & Mono./Bounded & 0 & --- \\
1 & sigmoid & $\sigma(x)$ & Mono./Bounded & 0 & --- \\
2 & relu & $\max(0, x)$ & Mono./Unbounded & 0 & --- \\
3 & identity & $x$ & Linear & 0 & --- \\
6 & lelu & $\max(0.01x, x)$ & Mono./Unbounded & 0 & --- \\
7 & softplus & $\ln(1 + e^x)$ & Mono./Unbounded & 0 & --- \\
9 & fs\_fast & $2\,\text{relu}(x)$ & Mono./Unbounded & 0 & --- \\
14 & gain\_mod & $x/(1{+}|x|)$ & Adaptive & 0 & --- \\
17 & integrate & $\tanh(x)(1{+}0.2\,e^{-|x|})$ & Mono./Bounded & 0 & --- \\
\bottomrule
\end{tabular}}
\end{table}

\subsection{Experimental Setup}
\label{sec:setup}

Experiments span logic problems (XOR, Parity-2 through Parity-8), classification (Two Spirals, concentric circles, step function), continuous regression (sine approximation), and multi-step reasoning (Tower of Hanoi). All use feedforward substrate topology unless noted, 30 independent replications (random seeds 42--71), and a 0.95 fitness threshold for ``solved'' (0.90 for concentric circles).

\paragraph{Reproducibility.} Each individual is a NEAT-evolved \cppn{} that emits connection weights and, in CPPN-output mode, one extra channel decoded to a per-node activation index. Candidate substrate nodes occupy a multi-resolution grid, where level $i$ contributes $4^{i+1}$ cells (e.g., 84 positions at depth~2, 21{,}844 at depth~6); CPPN-variance filtering (threshold 0.03) retains the informative ones. NEAT uses connection/node add--delete mutation (probabilities 0.5/0.5/0.2/0.2), compatibility threshold 3.0, genome elitism 15, survival threshold 0.2, and CPPN node activations drawn from \{tanh, sin, gauss\}. Unless noted, the population is 150 and each run is capped at 300 generations. The Population Size Invariance study instead sweeps the population over 50, 150, 500, 1{,}000, and 2{,}000; it and the Parity-scaling study use a 100-generation cap. Recurrent runs add hidden-to-hidden, backward, lateral, and self connections and iterate the substrate deterministically for $2^{d}{+}1$ steps at depth $d$. Complete configurations and result data are in the repository (Data/Code link above).

All group comparisons use Mann--Whitney $U$ tests (two-group) or Fisher's exact tests (proportions); correlations use Spearman's rank coefficient. Unless noted otherwise, reported differences are statistically significant at $p < 0.01$.

%% =============================================================================
%% SECTION 5: RESULTS
%% =============================================================================
\section{Results}
\label{sec:results}

\subsection{Per-Function Solvability}
\label{sec:solvability}

Parity-N problems (output 1 if an odd number of $N$ binary inputs are active) require alternating decision boundaries. Testing activation functions on Parity-4 reveals a three-tier solvability structure (Table~\ref{tab:palette}).

On XOR (simplest parity), sin converges 16$\times$ faster than a monotonic-only baseline palette (tanh, sigmoid, relu, identity): 1.1 vs 17.5 generations (Mann--Whitney $U{=}18$, $p{=}1.67{\times}10^{-11}$, rank-biserial $r{=}0.96$). Gaussian and sinusoidal activations are both smooth and bounded, yet sin achieves 100\% on parity while Gaussian achieves only 33\%. Both cross zero, but only sin does so \textit{periodically}. The decisive property is periodicity, not smoothness.

In evolved substrates with sparse hidden layers, the few hidden nodes produced by CPPN variance analysis fall short of the width-$\Theta(n)$ monotonic construction (Proposition~\ref{prop:monotonic}), which evolution does not discover. This divide is not universal (see the Discussion).

\subsection{Formal Analysis}
\label{sec:proof}

The oscillatory/monotonic divide we observe is a barrier of \textit{search and sparsity}, not of representation. We make both sides precise and locate the gap.

\begin{proposition}[Oscillatory sufficiency]
\label{prop:oscillatory}
A single hidden unit with cosine activation computes $n$-bit parity $P_n$ for any $n$. Let \mbox{$h(\mathbf{x}) = \cos\!\bigl(\pi \sum_{i=1}^n x_i\bigr)$}. For $\mathbf{x} \in \{0,1\}^n$, $\sum_i x_i = k \in \{0,\ldots,n\}$, so \mbox{$h(\mathbf{x}) = \cos(\pi k) = (-1)^k$}: $+1$ when $k$ is even and $-1$ when odd. The linear readout \mbox{$y = \tfrac{1}{2}(1 - h(\mathbf{x}))$} outputs $P_n$ with a single hidden unit, independent of $n$.
\end{proposition}

\begin{proposition}[Monotonic feasibility]
\label{prop:monotonic}
A single monotonic unit (perceptron) cannot compute $P_n$ for $n \geq 2$~\citep{minsky1969perceptrons}, but a single hidden layer of monotonic units can, with width $\Theta(n)$. The ``thermometer'' construction uses $n$ threshold units $h_i = \mathbf{1}\!\left[\sum_j x_j \geq i\right]$ and a sign-alternating readout $y = \sum_{i=1}^{n} (-1)^{i+1} h_i$; if $s = \sum_j x_j$ then $y = \sum_{i=1}^{s} (-1)^{i+1} = P_n(\mathbf{x})$. Parity is thus representable by monotonic networks of modest width.
\end{proposition}

These bounds concern representation. The phenomenon our experiments expose lies between representability and \textit{discoverability under evolutionary search in sparse substrates}. Variance-based substrate discovery yields few hidden nodes, so the width-$\Theta(n)$ monotonic construction of Proposition~\ref{prop:monotonic}, which also requires coordinating the alternating readout weights, is not found by neuroevolution, while a single oscillatory unit (Proposition~\ref{prop:oscillatory}) needs neither width nor coordination and is found within a few generations. Gradient descent with a fixed wide layer reaches the monotonic solution reliably (tanh solves Parity-8 with $h{=}32 < 2^{7}$ units; Table~\ref{tab:gradient_baseline}), confirming the barrier is search and sparsity, not representation.

\subsection{Per-Function Ablation}
\label{sec:ablation}

We test each of the 18 palette functions individually on Parity-4 using uniform activation mode (population 150, 300 generation maximum). Table~\ref{tab:palette} shows both function definitions and results.

\begin{figure}[htbp]
\centering
\includegraphics[width=\columnwidth]{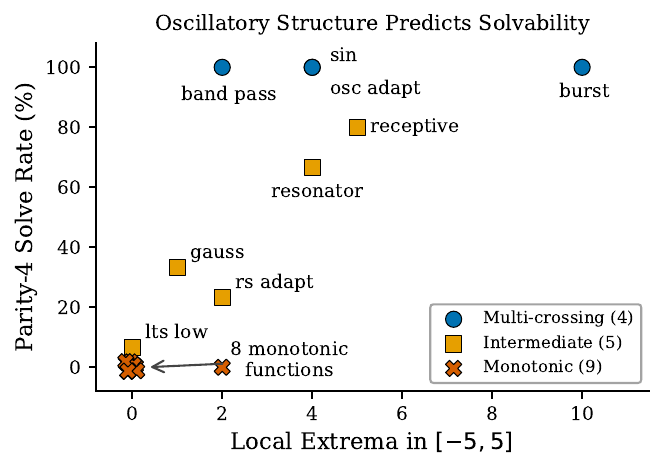}
\caption{Oscillatory structure predicts solvability. Local extrema count in $[-5, 5]$ vs Parity-4 solve rate ($N{=}30$) for all 18 activation functions. Functions with more local extrema achieve higher solve rates. The two outliers at 2 extrema reflect extrema amplitude: band\_pass has deep sign-reversing extrema (100\%), while integrate has shallow perturbations of a monotonic trend (0\%).}
\label{fig:zero_crossing}
\end{figure}

The solvability boundary follows oscillatory structure, quantified by local extrema count (Spearman $\rho{=}0.81$, $p{=}4.4{\times}10^{-5}$), not category label. Four functions achieve 100\%: sin (2.9 gens), osc\_adapt (3.7), burst (10.7), and band\_pass (69.2). Five functions form an intermediate tier. Rs\_adapt (23.3\%) has hidden oscillatory behavior: $\tanh(x)(1{-}0.3|x|)$ reverses sign at $|x| \approx 3.3$. Lts\_low (6.7\%), a shifted sigmoid, is genuinely monotonic, solving only by threshold proximity rather than true parity representation, consistent with Proposition~\ref{prop:monotonic}. All 9 remaining strictly monotonic functions achieve 0\%. However, monotonic functions are not helpless: their mean plateau fitness is $0.851 \pm 0.048$ (${\sim}$85\% of patterns correct on Parity-4, $N{=}270$ across 9 functions). Evolution gets close but cannot cross the 0.95 threshold, which requires resolving the alternating parity structure that monotonic units capture only with width (Proposition~\ref{prop:monotonic}). The divide is depth-invariant: sin achieves 100\% and tanh 0\% at depths 2, 4, and 6. A $260\times$ increase in available substrate capacity does not rescue monotonic functions, indicating an evolvability barrier rather than a capacity shortage, consistent with Proposition~\ref{prop:monotonic}, where monotonic parity needs width that variance-filtered substrates and evolutionary search do not supply. Measuring the evolved substrates directly confirms this: the sin networks that solve parity use a median of 15 hidden nodes and tanh's best-of-failed networks a comparable 10 hidden nodes, both far exceeding the few units the monotonic construction needs, yet tanh never solves. The connection counts agree: tanh's failed substrates carry a median of 44.5 connections (range 2--96), again more than the sparse monotonic construction requires, even though they stay below the 100-connection budget that most sin solutions reach. Figure~\ref{fig:zero_crossing} quantifies this: a gradient from undamped oscillations (100\%) to strictly monotonic (0\%).

\paragraph{Per-function testing on XOR.} Repeating the per-function test on XOR (300 gen max) preserves the three-tier structure with shifted thresholds: all 7 oscillatory functions solve 100\%, monotonic functions only partially (tanh 70\%, relu 40\%, sigmoid 26.7\%, identity 0\%). XOR is equivalent to Parity-2, and the higher tanh rate here (70\% vs 50\% in Table~\ref{tab:parity_scaling}) reflects the longer generation limit (300 vs 100) and XOR being the easiest parity instance; the added search budget mitigates but does not eliminate the monotonic barrier. Per-function solve rates correlate strongly across the two tasks (Spearman $\rho{=}0.84$), so the divide is not task-specific.

\subsection{Parity Scaling}
\label{sec:parity_scaling}

Using sin activation in uniform mode, we evolve independent populations on Parity-2 through Parity-8. Table~\ref{tab:parity_scaling} shows the results.

\begin{table}[htbp]
\centering
\caption{Parity scaling: sin vs tanh ($N{=}30$ seeds each; population 150; 100 gen max). Sin achieves 100\% across all sizes; tanh degrades monotonically. Percentages are solve rates (replications reaching the threshold).}
\label{tab:parity_scaling}
{\small\setlength{\tabcolsep}{3pt}
\begin{tabular}{lccccc}
\toprule
Problem & Samples & Sin\% & Sin Gen & Tanh\% & Tanh Fit. \\
\midrule
Parity-2 & 4 & 100 & 1.0 & 50.0 & 0.930 \\
Parity-3 & 8 & 100 & 2.2 & 3.3 & 0.890 \\
Parity-4 & 16 & 100 & 2.9 & 0 & 0.863 \\
Parity-5 & 32 & 100 & 2.9 & 0 & 0.826 \\
Parity-6 & 64 & 100 & 3.4 & 0 & 0.825 \\
Parity-7 & 128 & 100 & 3.7 & 0 & 0.821 \\
Parity-8 & 256 & 100 & 3.6 & 0 & 0.791 \\
\bottomrule
\end{tabular}}
\end{table}

\textit{Sin: 100\% across all sizes}, from 4 to 256 samples, within 1--4 generations on average, consistent with Proposition~\ref{prop:oscillatory}. \textit{Tanh: monotonic degradation}. Even on Parity-2, tanh achieves only 50\%; from Parity-3 onward, categorically 0\%. Mean best fitness degrades from 0.930 (Parity-2) to 0.791 (Parity-8), consistent with the width-and-coordination barrier of Proposition~\ref{prop:monotonic}. Convergence speed for sin is flat (1.0--3.7 gens across 64$\times$ sample range), suggesting the difficulty scales sublinearly with problem size.

\subsection{Population Size Invariance}
\label{sec:pop_invariance}

Larger populations explore more of the weight space, potentially discovering the width-$\Theta(n)$ monotonic construction (Proposition~\ref{prop:monotonic}). We re-evaluate sin vs tanh on Parity-4 with the population genuinely varied from 50 to 2{,}000 (Table~\ref{tab:pop_sensitivity}).

\begin{table}[htbp]
\centering
\caption{Population reshapes the dynamics but not the oscillatory divide (Parity-4, $N{=}30$ per cell, 100 gen max). Sin solves 100\% at every size; tanh 0\% at every size. Percentages are solve rates.}
\label{tab:pop_sensitivity}
{\small\setlength{\tabcolsep}{4pt}
\begin{tabular}{lcccc}
\toprule
Pop Size & Sin\% & Sin Gen & Tanh\% & Tanh Fit. \\
\midrule
50 & 100 & 5.0 & 0 & 0.808 \\
150 & 100 & 3.0 & 0 & 0.863 \\
500 & 100 & 2.0 & 0 & 0.894 \\
1{,}000 & 100 & 1.0 & 0 & 0.900 \\
2{,}000 & 100 & 1.0 & 0 & 0.910 \\
\bottomrule
\end{tabular}}
\end{table}

Across a 40$\times$ range, the solve-rate divide is invariant: sin 100\% and tanh 0\% at every population. Population is not inert, however. It \textit{accelerates} sin: median convergence falls from 5 generations (pop 50) to 1 (pop $\geq$1{,}000) as larger pools sample a working oscillatory unit sooner. It also lifts tanh's best-found fitness monotonically from 0.808 to 0.910, yet tanh never crosses the 0.95 threshold at any size. Added search budget brings monotonic networks closer to parity but supplies neither the width nor the weight coordination that Proposition~\ref{prop:monotonic} requires.

\subsection{Assignment Method Comparison}
\label{sec:assignment_comparison}

Having established the solvability divide on parity (RQ1), we turn to RQ2: which assignment mechanism best places per-node functions? We use a spatial problem rather than parity, since assignment matters most where function choice interacts with geometry. Two Spirals (194 interleaved points, binary classification) provides a non-trivial testbed for comparing the three assignment methods. Table~\ref{tab:assignment} shows results from 9 configurations tested at depth 3.

\begin{table}[htbp]
\centering
\caption{Assignment method comparison on Two Spirals ($N{=}30$; depth 3). CPPN-output significantly outperforms the other methods; weight-interpretation gives no gain over the uniform baseline. cppn\_output\_$k$ = CPPN-output over a $k$-function palette (4: monotonic only; 6: adds sin, gauss).}
\label{tab:assignment}
{\footnotesize\setlength{\tabcolsep}{2.5pt}
\begin{tabular}{llcc}
\toprule
Method & Configuration & Accuracy & Std \\
\midrule
\multirow{3}{*}{CPPN-output} & cppn\_output\_4 & \textbf{76.9\%} & 0.29 \\
& cppn\_output\_6 & \textbf{76.9\%} & 0.29 \\
& cppn\_with\_sin & 76.1\% & 0.42 \\
\midrule
\multirow{3}{*}{Weight-interp.} & sign-based & 75.2\% & 0.14 \\
& magnitude-based & 75.2\% & 0.14 \\
& variance-based & 75.2\% & 0.14 \\
\midrule
\multirow{3}{*}{Uniform} & disabled & 75.2\% & 0.14 \\
& global\_sin & 75.2\% & 0.14 \\
& global\_tanh & 75.2\% & 0.14 \\
\bottomrule
\end{tabular}}
\end{table}

Weight-interpretation provides zero benefit over no function selection: all three variants collapse to the 75.2\% uniform baseline. CPPN-controlled assignment yields only a marginal gain on Two Spirals (76.9\% vs 75.2\%), near the feedforward ceiling. The decisive evidence is on concentric circles, where the effect is categorical: CPPN-output with the 6-function palette achieves 66.7\% vs 0\% for both weight-interpretation and uniform methods (Fisher's exact $p{=}1.4{\times}10^{-8}$). Adding sin to the CPPN palette \textit{hurts} Two Spirals (76.1\% vs 76.9\%), consistent with substrate matching: oscillatory functions are mismatched to spatial classification.

\subsection{Palette Size Robustness}
\label{sec:palette_robustness}

How sensitive is performance to palette size? Table~\ref{tab:antagonism} shows that all palette sizes from 1 to 13 of the 18 available functions achieve 100\% success rate on Parity-4. The search space expansion argument ($k^N$ configurations) predicts exponential degradation; success rate is unaffected, though convergence slows 5.5$\times$ from 1 to 13 functions (1.7$\rightarrow$9.3 generations, with standard deviation increasing from 0.7 to 6.4). 
\begin{table}[htbp]
\centering
\caption{Palette size vs convergence speed ($N{=}30$ seeds per size; Parity-4; all 100\% success).}
\label{tab:antagonism}
{\small\setlength{\tabcolsep}{4pt}
\begin{tabular}{lccc}
\toprule
Palette Size & Functions & Avg Gen & Std \\
\midrule
1 (sin only) & sin & 1.7 & 0.7 \\
3 & sin, tanh, relu & 3.4 & 1.2 \\
4 & +sigmoid & 3.1 & 1.1 \\
5 & +identity & 3.2 & 1.2 \\
6 & +gauss & 2.6 & 1.1 \\
7 & +lelu & 3.3 & 1.0 \\
8 & +softplus & 3.4 & 1.2 \\
13 & 13 functions & 9.3 & 6.4 \\
\bottomrule
\end{tabular}}
\end{table}

\subsection{Recurrence Collapses the Divide}
\label{sec:recurrence}

Table~\ref{tab:ff_vs_rec} compares each function's solve rate under feedforward versus recurrent topology on Parity-4 (threshold 0.95). Recurrence partially collapses the divide: all oscillatory and quasi-oscillatory functions remain at 100\%, and lts\_low rises from 6.7\%$\rightarrow$100\%. Among monotonic functions, the collapse is graded: softplus 0\%$\rightarrow$90\%, sigmoid 0\%$\rightarrow$70\%, gauss 33\%$\rightarrow$77\%, but tanh remains near 0\% (6.7\%) and identity, lelu stay at 0\%. Recurrence gives bounded monotonic functions temporal iteration sufficient for some seeds, but unbounded functions (relu 3.3\%, fs\_fast 6.7\%, identity 0\%, lelu 0\%) cannot benefit: their unbounded self-application diverges rather than producing oscillatory dynamics.

\begin{table}[htbp]
\centering
\caption{Feedforward vs recurrent topology on Parity-4 (both $N{=}30$, 300 gen max, pop 150; solve threshold 0.95). Recurrence partially collapses the divide: oscillatory functions remain at 100\%; bounded monotonic partially solve; unbounded/linear remain near 0\%. \% = solve rate.}
\label{tab:ff_vs_rec}
{\footnotesize\setlength{\tabcolsep}{2pt}
\begin{tabular}{lcccc}
\toprule
Function & FF\% & FF Gen & Rec\% & Rec Gen \\
\midrule
sin & 100 & 2.9 & 100 & 2.0 \\
osc\_adapt & 100 & 3.7 & 100 & 3.0 \\
burst & 100 & 10.7 & 100 & 5.0 \\
band\_pass & 100 & 69.2 & 100 & 10.0 \\
\midrule
receptive & 80 & 100.1 & \textbf{100} & 3.0 \\
resonator & 66.7 & 55.9 & \textbf{100} & 2.0 \\
gauss & 33.3 & 139.3 & \textbf{76.7} & 95.0 \\
rs\_adapt & 23.3 & 202.4 & \textbf{30.0} & 165.0 \\
lts\_low & 6.7 & 246.0 & \textbf{100} & 4.0 \\
\midrule
softplus & 0 & --- & \textbf{90.0} & 7.0 \\
sigmoid & 0 & --- & \textbf{70.0} & 7.0 \\
integrate & 0 & --- & \textbf{30.0} & 90.0 \\
gain\_mod & 0 & --- & \textbf{13.3} & 6.5 \\
tanh & 0 & --- & 6.7 & 98.5 \\
fs\_fast & 0 & --- & 6.7 & 210.0 \\
relu & 0 & --- & 3.3 & 271.0 \\
identity & 0 & --- & 0 & --- \\
lelu & 0 & --- & 0 & --- \\
\bottomrule
\end{tabular}}
\end{table}

\paragraph{Parity-8 with recurrence.}
Under full recurrence on Parity-8, all 4 oscillatory functions still solve 100\%, and the monotonic collapse stays category-dependent: bounded/radial functions partially solve (32.2\%; sigmoid 70\%, softplus 50\%, gauss 27\%) while unbounded/linear remain at 0\%.

\paragraph{Which connection types matter?}
Table~\ref{tab:recurrence_type} decomposes full recurrence by connection type (tanh on Parity-4, 0.95 threshold). Hidden-to-hidden connections alone do most of the work (20\% vs 0\% feedforward), while backward, lateral, and self-loops add little (6.7--13.3\%).

\begin{table}[htbp]
\centering
\caption{Connection type decomposition (tanh on Parity-4, $N{=}30$, 300 gen max, pop 150; solve threshold 0.95). Each row adds one connection type to hidden-to-hidden (h-to-h) as base.}
\label{tab:recurrence_type}
{\small\setlength{\tabcolsep}{4pt}
\begin{tabular}{lccc}
\toprule
Connection Type & Solve\% & Mean Gen & Avg Fitness \\
\midrule
feedforward only & 0 & --- & 0.914 \\
h-to-h only & \textbf{20.0} & 79.5 & 0.911 \\
h-to-h + backward & 13.3 & 191.2 & 0.917 \\
h-to-h + lateral & 6.7 & 117.5 & 0.881 \\
h-to-h + self-loops & 6.7 & 58.5 & 0.885 \\
all recurrent & 6.7 & 98.5 & 0.893 \\
\bottomrule
\end{tabular}}
\end{table}

\paragraph{Ceilings persist under recurrence.}
Despite recurrence partially collapsing the solvability divide on parity, feedforward performance ceilings survive elsewhere. Across 240 recurrent runs on Tower of Hanoi, 0\% solve. On Two Spirals, 0\% across 240 runs. On concentric circles, recurrence has the \textit{opposite} effect across all 8 tested functions (Table~\ref{tab:circles_recurrence}; in both full-recurrence and hidden-only modes).

\begin{table}[htbp]
\centering
\caption{Recurrence on concentric circles ($N{=}30$ each; solve threshold 0.90). Opposite to parity, recurrence \textit{degrades} most functions; only osc\_adapt stays at 100\%. FF\% = feedforward solve rate; Full Rec\% = full recurrence; Hidden-only\% = recurrence on hidden-to-hidden connections only.}
\label{tab:circles_recurrence}
{\footnotesize\setlength{\tabcolsep}{3pt}
\begin{tabular}{lccc}
\toprule
Function & FF\% & Full Rec\% & Hidden-only\% \\
\midrule
osc\_adapt & 100 & \textbf{100} & \textbf{100} \\
sin & 100 & 80 & 50 \\
resonator & 100 & 70 & 57 \\
gauss & 43 & 3.3 & 3.3 \\
burst & 0 & 0 & 0 \\
sigmoid & 3.3 & 0 & 6.7 \\
softplus & 0 & 0 & 6.7 \\
tanh & 0 & 0 & 0 \\
\bottomrule
\end{tabular}}
\end{table}

Osc\_adapt alone is immune to recurrence-induced degradation on circles: its damped oscillation provides spatial locality that survives recurrent iteration, while pure sin (80\%) and resonator (70\%) degrade, and monotonic functions stay near 0\%. Substrate matching thus needs \textit{architecture--problem} alignment too: temporal iteration helps alternating boundaries (parity) but disrupts the spatial coherence that radial enclosure (circles) needs.

\subsection{Non-Parity Generalization}
\label{sec:nonparity}

Does the divide generalize beyond parity? We test all 18 functions on three non-parity problems: sine regression (continuous periodic), step function (linear classification), and concentric circles (radial classification). Table~\ref{tab:nonparity} summarizes results for representative functions across all four problem types.

\begin{table}[htbp]
\centering
\caption{Cross-problem comparison for representative functions. The three-tier solvability structure replicates on problems requiring non-monotonic decision surfaces (Parity-4, Circles) and vanishes on smooth problems (Sine, Step). All cells $N{=}30$. \% = solve rate; Gen = median generations to solve.}
\label{tab:nonparity}
{\footnotesize\setlength{\tabcolsep}{2pt}
\begin{tabular}{lcccccccc}
\toprule
& \multicolumn{2}{c}{Parity-4} & \multicolumn{2}{c}{Sine Regr.} & \multicolumn{2}{c}{Step Func.} & \multicolumn{2}{c}{Circles} \\
\cmidrule(lr){2-3} \cmidrule(lr){4-5} \cmidrule(lr){6-7} \cmidrule(lr){8-9}
Function & \% & Gen & \% & Gen & \% & Gen & \% & Gen \\
\midrule
sin & 100 & 2.9 & 100 & 1 & 97 & 55 & 100 & 23 \\
osc\_adapt & 100 & 3.7 & 100 & 1 & 100 & 49 & 100 & 2 \\
burst & 100 & 10.7 & 100 & 6 & 100 & 24 & 0 & --- \\
\midrule
resonator & 67 & 56 & 100 & 1 & 100 & 50 & 100 & 8 \\
gauss & 33 & 139 & 100 & 2 & 13 & 152 & 43 & 155 \\
rs\_adapt & 23 & 202 & 100 & 1 & 97 & 52 & 47 & 149 \\
\midrule
tanh & 0 & --- & 100 & 1 & 100 & 18 & 0 & --- \\
sigmoid & 0 & --- & 100 & 10 & 100 & 38 & 3.3 & 121 \\
identity & 0 & --- & 100 & 2 & 100 & 4 & 0 & --- \\
relu & 0 & --- & 100 & 6 & 83 & 14 & 3.3 & 271 \\
\bottomrule
\end{tabular}}
\end{table}

\paragraph{Sine regression.} All 18 functions achieve 100\%. The divide vanishes when smooth interpolation replaces discrete boundary partitioning.

\paragraph{Concentric circles.} The three-tier structure replicates. Radial boundaries, like parity, require non-monotonic decision surfaces: oscillatory functions solve (osc\_adapt 100\%, sin 100\%), while most monotonic functions fail (7 of 9 at 0\%; sigmoid and relu reach 3.3\%).

\paragraph{Step function.} The pattern inverts. Identity converges fastest (median 4 gens). Monotonic functions outperform oscillatory ones (sin median 55 gens), while localized functions like gauss struggle (13\%) on this global linear boundary. Oscillatory structure is counterproductive when the problem is linearly separable.

The divide emerges wherever problems require non-monotonic decision surfaces and disappears on smooth problems: match function complexity to \textit{decision boundary complexity}.

%% =============================================================================
%% SECTION 6: DISCUSSION
%% =============================================================================
\section{Discussion}
\label{sec:discussion}

\subsection{The Divide Is Optimizer-Specific}
\label{sec:gradient_baseline}

Is the oscillatory/monotonic divide universal or specific to evolutionary search? We trained standard MLPs (single hidden layer with $h$ hidden units, Adam optimizer, $lr{=}0.01$, 5{,}000 epochs max) with sin vs tanh activation on parity tasks at multiple values of $h$ (Table~\ref{tab:gradient_baseline}).

\begin{table}[htbp]
\centering
\caption{Gradient-trained MLP baseline (Adam, $lr{=}0.01$, 5{,}000 epochs max, $N{=}30$). The divide progressively reverses: tanh maintains 100\% through Parity-8; sin degrades to 0\%. Increasing $h$ does not help sin; it makes it worse on Parity-6 (66.7\%$\rightarrow$46.7\%).}
\label{tab:gradient_baseline}
{\small\setlength{\tabcolsep}{3pt}
\begin{tabular}{llccc}
\toprule
Problem & Act. & $h$ & Solve\% & Avg Epoch \\
\midrule
Parity-4 & sin & 16 & 100 & 257 \\
Parity-4 & tanh & 16 & 100 & 182 \\
\midrule
Parity-5 & sin & 16 & 100 & 477 \\
Parity-5 & sin & 32 & 100 & 543 \\
Parity-5 & tanh & 16 & 100 & 245 \\
\midrule
Parity-6 & sin & 16 & 66.7 & 2{,}653 \\
Parity-6 & sin & 32 & 46.7 & 2{,}986 \\
Parity-6 & tanh & 16 & 100 & 352 \\
Parity-6 & tanh & 32 & 100 & 370 \\
\midrule
Parity-7 & sin & 16 & 36.7 & 3{,}805 \\
Parity-7 & sin & 32 & 10.0 & 3{,}217 \\
Parity-7 & tanh & 16 & 100 & 563 \\
Parity-7 & tanh & 32 & 100 & 572 \\
\midrule
Parity-8 & sin & 32 & 0 & --- \\
Parity-8 & sin & 64 & 0 & --- \\
Parity-8 & tanh & 32 & 100 & 1{,}078 \\
\bottomrule
\end{tabular}}
\end{table}

On Parity-4 and Parity-5, both activations achieve 100\% at all tested $h$, with tanh converging faster.

From Parity-6 onward, the divide progressively reverses. Sin degrades from 100\% (Parity-5) to 47--67\% (Parity-6), 10--37\% (Parity-7), and 0\% (Parity-8, including $h{=}64$), while tanh stays at 100\%. Counterintuitively, \textit{more} hidden units hurt sin (Parity-6: 66.7\% at $h{=}16$ vs 46.7\% at $h{=}32$): Xavier initialization~\citep{glorot2010understanding} concentrates pre-activations near zero where $\sin(x) \approx x$, and small gradient steps never leave that quasi-linear regime~\citep{sitzmann2020siren,ziyin2020neural}. Indirect encoding sidesteps this: CPPN-generated and NEAT-mutated weights span the full range, so sin starts in its oscillatory regime, while gradient descent coordinates the $\Theta(n)$ monotonic units of Proposition~\ref{prop:monotonic} that evolutionary search cannot.

\subsection{Practical Implications}

Our results yield concrete guidelines for neuroevolution with indirect encoding:

\paragraph{Function-problem matching.} If a problem requires multi-crossing decision boundaries (parity, XOR chains, alternating patterns), include oscillatory functions, since monotonic-only palettes will fail under evolutionary search. For smooth regression or linear classification, function category is irrelevant: all 18 functions solve sine regression at 100\%. Among the 18 functions, osc\_adapt alone maintains 100\% across all tested conditions (parity, circles, feedforward, recurrent), making it a safe default.

\paragraph{Assignment method.} CPPN-output assignment outperforms weight-interpretation and uniform methods. Add an extra CPPN output for function selection rather than deriving it from weights.

\paragraph{Palette sizing.} Larger palettes slow convergence (5.5$\times$ from 1 to 13 functions on Parity-4) but do not reduce success rate. When problem structure is unknown, use the full palette. Slower convergence is acceptable insurance against choosing the wrong category.

\subsection{Limitations}

\paragraph{Task scope.} Our evidence spans low-dimensional problems ($\leq$8 inputs). High-dimensional tasks (vision, robotics) may blur the divide. Prior work on ES-HyperNEAT achieved 29\% on MNIST~\citep{claret2024eshyperneat}; even partitioning the input into spatially specialized experts reaches only 49\%~\citep{claret2026partitioned}, far below the 98\%+ of gradient methods.

\paragraph{Representational scope.} Our feedforward results reflect evolved substrates with few hidden nodes; monotonic functions can represent parity given sufficient width~(Proposition~\ref{prop:monotonic}), but neuroevolution did not discover that construction in our experiments.

\paragraph{Architectural constraints.} The Hanoi plateau ($\sim$78\%) and Two Spirals ceiling ($\sim$77\%) persist under both feedforward and recurrent architectures, indicating deeper problem--architecture mismatches that topology alone cannot resolve.

\paragraph{Biological parallels.} The evolved functional heterogeneity mirrors a cortical principle: distinct neuron types coexist because different computational demands require different response properties, not because one type is universally better. Uniform activation is to a network what a single neuron type would be to cortex: sufficient for simple tasks, inadequate when the problem needs qualitatively different primitives. Whether evolution organizes these per-node assignments into spatially coherent functional regions is left to future work.

%% =============================================================================
%% SECTION 7: CONCLUSION
%% =============================================================================
\section{Conclusion}
\label{sec:conclusion}

We addressed three research questions about per-node activation function evolution in indirect encoding.

\paragraph{RQ1: Categorical solvability limits.} Activation selection is substrate matching: the per-node assignment is evolved, not a tuned hyperparameter, and function category is a categorical determinant of solvability, with oscillatory functions solving parity at 100\% where monotonic functions reach 0\%. Both sides are formal (Propositions~\ref{prop:oscillatory},~\ref{prop:monotonic}): monotonic networks need width $\Theta(n)$ that neuroevolution did not discover in our sparse-substrate experiments, while one oscillatory unit suffices. The 18-function divide is predicted by oscillatory structure ($\rho{=}0.81$) and holds across population sizes 50--2{,}000.

\paragraph{RQ2: Assignment mechanism.} CPPN-controlled per-node assignment significantly outperforms weight-interpretation and uniform methods.

\paragraph{RQ3: Boundary conditions.} The divide is \textit{problem-specific} (absent on sine regression, present on concentric circles), \textit{architecture-dependent} (recurrence collapses it on parity but \textit{widens} it on circles), and \textit{optimizer-specific} (gradient descent inverts it: tanh solves through Parity-8 while sin degrades to 0\%).

Per-node activation selection in indirect encoding lets evolution search a design space impractical by hand, shaping substrates through function category, problem geometry, and optimizer.

%% =============================================================================
%% ACKNOWLEDGMENTS & REFERENCES
%% =============================================================================
\section{Acknowledgments}
Generative AI tools assisted with visualizations and draft revision.

\footnotesize
\bibliographystyle{apalike}
\bibliography{references}

@article{stanley2002evolving,
  title={Evolving neural networks through augmenting topologies},
  author={Stanley, Kenneth O and Miikkulainen, Risto},
  journal={Evolutionary computation},
  volume={10},
  number={2},
  pages={99--127},
  year={2002},
  publisher={MIT Press}
}

@article{stanley2009hypercube,
  title={A hypercube-based encoding for evolving large-scale neural networks},
  author={Stanley, Kenneth O and D'Ambrosio, David B and Gauci, Jason},
  journal={Artificial life},
  volume={15},
  number={2},
  pages={185--212},
  year={2009},
  publisher={MIT Press One Rogers Street, Cambridge, MA 02142-1209, USA journals-info~…}
}

@article{risi2012enhanced,
  title={An enhanced hypercube-based encoding for evolving the placement, density, and connectivity of neurons},
  author={Risi, Sebastian and Stanley, Kenneth O},
  journal={Artificial life},
  volume={18},
  number={4},
  pages={331--363},
  year={2012},
  publisher={MIT Press 55 Hayward Street, Cambridge, MA 02142-1315, USA}
}

@inproceedings{hagg2017evolving,
  title={Evolving parsimonious networks by mixing activation functions},
  author={Hagg, Alexander and Mensing, Maximilian and Asteroth, Alexander},
  booktitle={Proceedings of the genetic and evolutionary computation conference},
  pages={425--432},
  year={2017}
}

@article{papavasileiou2021systematic,
  title={A systematic literature review of the successors of “neuroevolution of augmenting topologies”},
  author={Papavasileiou, Evgenia and Cornelis, Jan and Jansen, Bart},
  journal={Evolutionary computation},
  volume={29},
  number={1},
  pages={1--73},
  year={2021},
  publisher={MIT Press}
}

@inproceedings{papavasileiou2022hafdneat,
  title={Towards optimizing neural networks' connectivity and architecture simultaneously with feature selection},
  author={Papavasileiou, Evgenia and Cornelis, Jan P and Jansen, Bart},
  booktitle={Proceedings of the Genetic and Evolutionary Computation Conference Companion},
  pages={667--670},
  year={2022}
}

@article{ramachandran2017searching,
  title={Searching for activation functions},
  author={Ramachandran, Prajit and Zoph, Barret and Le, Quoc V},
  journal={arXiv preprint arXiv:1710.05941},
  year={2017}
}

@article{markram2004interneurons,
  title={Interneurons of the neocortical inhibitory system},
  author={Markram, Henry and Toledo-Rodriguez, Maria and Wang, Yun and Gupta, Anirudh and Silberberg, Gilad and Wu, Caizhi},
  journal={Nature reviews neuroscience},
  volume={5},
  number={10},
  pages={793--807},
  year={2004},
  publisher={Nature Publishing Group UK London}
}

@article{izhikevich2003simple,
  title={Simple model of spiking neurons},
  author={Izhikevich, Eugene M},
  journal={IEEE Transactions on neural networks},
  volume={14},
  number={6},
  pages={1569--1572},
  year={2003},
  publisher={IEEE}
}

@article{stanley2007compositional,
  title={Compositional pattern producing networks: A novel abstraction of development},
  author={Stanley, Kenneth O},
  journal={Genetic programming and evolvable machines},
  volume={8},
  number={2},
  pages={131--162},
  year={2007},
  publisher={Springer}
}

@inproceedings{gauci2008case,
  title={A case study on the critical role of geometric regularity in machine learning.},
  author={Gauci, Jason and Stanley, Kenneth O and others},
  booktitle={AAAI},
  pages={628--633},
  year={2008}
}

@article{zoph2016neural,
  title={Neural architecture search with reinforcement learning},
  author={Zoph, Barret and Le, Quoc V},
  journal={arXiv preprint arXiv:1611.01578},
  year={2016}
}

@inproceedings{real2019regularized,
  title={Regularized evolution for image classifier architecture search},
  author={Real, Esteban and Aggarwal, Alok and Huang, Yanping and Le, Quoc V},
  booktitle={Proceedings of the aaai conference on artificial intelligence},
  volume={33},
  number={01},
  pages={4780--4789},
  year={2019}
}

@article{lecun2015deep,
  title={Deep learning},
  author={LeCun, Yann and Bengio, Yoshua and Hinton, Geoffrey},
  journal={nature},
  volume={521},
  number={7553},
  pages={436--444},
  year={2015},
  publisher={Nature Publishing Group UK London}
}

@inproceedings{glorot2010understanding,
  title={Understanding the difficulty of training deep feedforward neural networks},
  author={Glorot, Xavier and Bengio, Yoshua},
  booktitle={Proceedings of the thirteenth international conference on artificial intelligence and statistics},
  pages={249--256},
  year={2010},
  organization={JMLR Workshop and Conference Proceedings}
}

@article{bingham2022pangaea,
  title={Discovering parametric activation functions},
  author={Bingham, Garrett and Miikkulainen, Risto},
  journal={Neural Networks},
  volume={148},
  pages={48--65},
  year={2022},
  publisher={Elsevier}
}

@article{nader2021evolution,
  title={Evolution of activation functions: an empirical investigation},
  author={Nader, Andrew and Azar, Danielle},
  journal={ACM Transactions on Evolutionary Learning and Optimization},
  volume={1},
  number={2},
  pages={1--36},
  year={2021},
  publisher={ACM New York, NY}
}

@article{sitzmann2020siren,
  title={Implicit neural representations with periodic activation functions},
  author={Sitzmann, Vincent and Martel, Julien and Bergman, Alexander and Lindell, David and Wetzstein, Gordon},
  journal={Advances in neural information processing systems},
  volume={33},
  pages={7462--7473},
  year={2020}
}

@incollection{miikkulainen2024evolving,
  title={Evolving deep neural networks},
  author={Miikkulainen, Risto and Liang, Jason and Meyerson, Elliot and Rawal, Aditya and Fink, Dan and Francon, Olivier and Raju, Bala and Shahrzad, Hormoz and Navruzyan, Arshak and Duffy, Nigel and others},
  booktitle={Artificial intelligence in the age of neural networks and brain computing},
  pages={269--287},
  year={2024},
  publisher={Elsevier}
}

@article{ziyin2020neural,
  title={Neural networks fail to learn periodic functions and how to fix it},
  author={Ziyin, Liu and Hartwig, Tilman and Ueda, Masahito},
  journal={Advances in Neural Information Processing Systems},
  volume={33},
  pages={1583--1594},
  year={2020}
}

@article{noel2021gcu,
  title={Growing cosine unit: A novel oscillatory activation function that can speedup training and reduce parameters in convolutional neural networks},
  author={Noel, Mathew Mithra and Trivedi, Advait and Dutta, Praneet and others},
  journal={arXiv preprint arXiv:2108.12943},
  year={2021}
}

@inproceedings{rahaman2019spectral,
  title={On the spectral bias of neural networks},
  author={Rahaman, Nasim and Baratin, Aristide and Arpit, Devansh and Draxler, Felix and Lin, Min and Hamprecht, Fred and Bengio, Yoshua and Courville, Aaron},
  booktitle={International conference on machine learning},
  pages={5301--5310},
  year={2019},
  organization={PMLR}
}

@inproceedings{sopena1999periodic,
  title={Neural networks with periodic and monotonic activation functions: a comparative study in classification problems},
  author={Sopena, Josep M and Romero, Enrique and Alquezar, Rene},
  booktitle={9th International Conference on Artificial Neural Networks: ICANN'99},
  pages={323--328},
  year={1999},
  organization={IET}
}

@book{minsky1969perceptrons,
    author = {Minsky, Marvin and Papert, Seymour A.},
    title = {Perceptrons: An Introduction to Computational Geometry},
    publisher = {The MIT Press},
    year = {2017},
    month = {09},
    isbn = {9780262343930},
    doi = {10.7551/mitpress/11301.001.0001},
    url = {https://doi.org/10.7551/mitpress/11301.001.0001},
}

@inproceedings{claret2026emr,
  author={Claret, Romain and O'Neill, Michael and Cotofrei, Paul and Stoffel, Kilian},
  title={Tensor-Accelerated Eager Multi-Resolution Grids for Evolving Large-Scale Substrates},
  booktitle={Proceedings of the Genetic and Evolutionary Computation Conference Companion (GECCO Companion '26)},
  year={2026},
  address={San Jose, Costa Rica},
  publisher={ACM},
  doi={10.1145/3795101.3805361}
}

@inproceedings{claret2024eshyperneat,
  author={Claret, Romain and O'Neill, Michael and Cotofrei, Paul and Stoffel, Kilian},
  title={Investigating Hyperparameter Optimization and Transferability for {ES-HyperNEAT}: A {TPE} Approach},
  booktitle={Proceedings of the Genetic and Evolutionary Computation Conference Companion (GECCO '24 Companion)},
  year={2024},
  publisher={ACM},
  pages={1879--1887},
  doi={10.1145/3638530.3664144}
}

@inproceedings{claret2026partitioned,
  title={Breaking the Central Bias: Spatially Partitioned Experts for Coordinate-Based Neuroevolution},
  author={Claret, Romain and Gygax, Arthur and O'Neill, Michael and Cotofrei, Paul and Felber, Pascal},
  booktitle={International conference on pattern recognition},
  year={2026},
  publisher={Springer}
}

\end{document}